\documentclass{article}

\usepackage{microtype}
\usepackage{subfigure}
\usepackage{booktabs} % for professional tables

\usepackage[preprint]{neurips_2023}
\usepackage[pdftex]{graphicx}
\usepackage{wrapfig}

\usepackage[utf8]{inputenc} % allow utf-8 input
\usepackage[T1]{fontenc}    % use 8-bit T1 fonts
\usepackage{hyperref}       % hyperlinks
\usepackage{url}            % simple URL typesetting
\usepackage{booktabs}       % professional-quality tables
\usepackage{amsfonts}       % blackboard math symbols
\usepackage{nicefrac}       % compact symbols for 1/2, etc.
\usepackage{microtype}      % microtypography
\usepackage{xcolor}         % colors

\usepackage{amsmath}
\usepackage{amssymb}
\usepackage{mathtools}
\usepackage{amsthm}

\title{Hierarchical reinforcement learning with natural language subgoals}

\author{%
Arun Ahuja \\
DeepMind New York City \\
\And
Kavya Kopparapu\\
DeepMind New York City \\
\And
Rob Fergus\\
DeepMind New York City \\
\And
Ishita Dasgupta\thanks{idg@deepmind.com}\\
DeepMind New York City
}

\begin{document}

\maketitle

\begin{abstract}
Hierarchical reinforcement learning has been a compelling approach for achieving goal directed behavior over long sequences of actions. However, it has been challenging to implement in realistic or open-ended environments. A main challenge has been to find the right space of sub-goals over which to instantiate a hierarchy. We present a novel approach where we use data from humans solving these tasks to softly supervise the goal space for a set of long range tasks in a 3D embodied environment. In particular, we use unconstrained natural language to parameterize this space. This has two advantages: first, it is easy to generate this data from naive human participants; second, it is flexible enough to represent a vast range of sub-goals in human-relevant tasks. Our approach outperforms agents that clone expert behavior on these tasks, as well as HRL from scratch without this supervised sub-goal space. Our work presents a novel approach to combining human expert supervision with the benefits and flexibility of reinforcement learning.

\end{abstract}

\section{Introduction}

Despite several recent successes of reinforcement learning, a major challenge has been using it in real world settings. Goal-directed behavior over long time horizons has thus far been challenging for traditional RL and its relatively data hungry process of exploration and temporal credit assignment. This has been especially limiting in real-world-like embodied tasks that operate over motor-control action spaces that make even relatively simple tasks require a long series of motor actions. RL has primarily thrived in worlds that accommodate simple abstract action spaces like games, where a single `action' can elicit large changes in the environment. However, this is limiting -- a central advantage of generic embodied action spaces is that they are realistic, flexible, and permit open-ended and emergent behaviors. RL's inability to operate over these action spaces (due to challenges in exploration and long-term credit assignment over long action sequences) has been a major impediment to its application in real-world settings.

An influential approach to extending RL to long range tasks has been to use hierarchies over the space of actions. Intuitively, this means that the `action space' that one actually does credit assignment and exploration over are temporally extended \textit{sequences of actions} \citep{sutton1998intra, hauskrecht2013hierarchical} that achieve \textit{subgoals} on the path toward achieving the target task. The main challenge here has been to devise (or learn) a general enough space of subgoals that both effectively reduces the planning horizon but is also expressive enough to permit interesting behaviors \citep{da2012learning, mankowitz2018learning}. The core challenge here is to find the right set of abstractions for a given domain and set of tasks.

In this work, we investigate natural language as a way to parameterize this subgoal space. Language is a lossy channel -- a text description of an agent trajectory will discard a lot of (detailed, grounded, visual) information. However, language has evolved explicitly to still be expressive enough to represent the vast majority of ideas, goals, and behaviors relevant to humans. This makes it a strong contender for specifying subgoals that both effectively reduce complexity, while retaining expressivity where it matters. 

Language also has the added advantage that we can crowd-source it from naive human participants. In this work, we explicitly elicit hierarchical trajectories with linguistic subgoals from human participants. One participant breaks down a task into sub-goals and another executes these sub-goals in an embodied action space. In this paper, we describe a way to use this data to softly supervise a hierarchical agent that can learn to solve complex long-horizon tasks in a 3-D embodied environment.

\begin{figure*}
    \centering
    \includegraphics[width=1.0\linewidth]{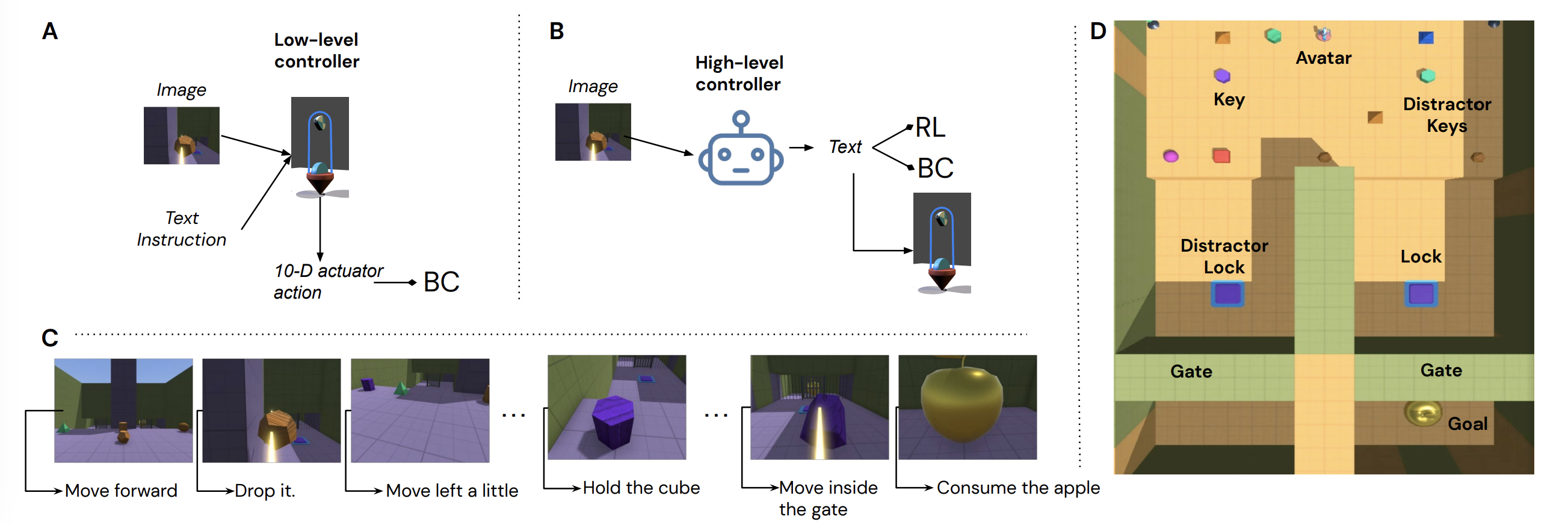}
    \caption{\textbf{Set up and example episode.} A. Observations, outputs and losses for the 'low-level' (LL) agent; B. Same for the 'high-level' (HL) agent. C. Example episode with observations to and output text from the HL agent; D. Top-down view of level Key Choice Hard.}
    \label{fig:schematic}
\end{figure*}

\section{Methods}
\subsection{Environment and tasks}
We use a 3-D embodied environment in Unity \citep{ward2020using}, showing proof of concept of our method on four tasks. Similar to the tasks described in DMLab \citep{beattie2016deepmind}, the goal in these tasks is to find and consume an apple. To acquire the goal apple, the agent must unlock a gate by placing a color-matched key object on a corresponding sensor (Fig \ref{fig:schematic}D). The main challenge in these tasks comes from requiring several steps, including information gathering (to know which key is needed and which sensor to place it on); details are in Appendix \ref{sec:task_details}. For the purpose of the main results, we classify the tasks into two Easy and two Hard tasks. The main feature of the Hard tasks that makes them more difficult than the Easy ones is that they contain several distractors, so finding the right key requires information gathering / exploration, and specifying which object to pick up can be challenging.

\subsection{Data collection}
Similar to \cite{abramson2020imitating} we collect data using two players, a `Setter' and a `Solver'. For the given tasks, a single controllable avatar is available and controlled by the `Solver'. Given the task goal, the `Setter' instructs the `Solver', via a chat interface, on how to solve the task. The `Setter' can observe the `Solver' but cannot interact with the environment directly. Data was collected across many goal-directed tasks, including those described here. 

\subsection{Agent training and architecture}
Our hierarchical agent has two components -  a `low-level' (LL) agent that produces motor commands for the agent and a `high-level' (HL) agent that provides subgoals for the agent. Both use the same architecture, as described in \cite{abramson2020imitating}.

\textbf{Low-level agent:} We first pre-train the `low-level' agent to follow relatively simple language commands (Fig \ref{fig:schematic}A). To do this we follow \cite{abramson2020imitating}, to imitate (i.e. behaviorally clone) expert humans on a large range of language conditional tasks in 3-D Unity environments similar to our goal environment. 

In particular, we use the `Solver' data - the player that receives language instructions and controls the embodied avatar to achieve these instructions -- to train this LL agent. This data includes observation sequences $\mathbf{o}_{\leq T} \equiv (\mathbf{o}_0, \mathbf{o}_1, \mathbf{o}_2, \dots, \mathbf{o}_T)$ (first person images from the environment), action sequences $\mathbf{a}_{\leq T} \equiv (\mathbf{a}_0, \mathbf{a}_1, \mathbf{a}_2, \dots, \mathbf{a}_T)$ (10-D actuator actions) and language instructions $\mathbf{g}_{\leq T} \equiv (\mathbf{g}_0, \mathbf{g}_1, \mathbf{g}_2, \dots, \mathbf{g}_T)$ (natural language sentences padded to length 24). With this data, we learn a policy that optimizes the supervised training objective:

\begin{align*}
\mathcal{L}^\textsc{LL}_{BC} & = -\frac{1}{B} \sum_{n=1}^{B} \sum_{t=0}^{K} \ln \pi(\mathbf{a}_{n,t} \mid \mathbf{o}_{n,\leq t}, \mathbf{g}_{n,t})
\end{align*} 
where $B$ is the minibatch size, $K$ is the backpropagation-through-time window size. 

The LL agent is then frozen. Our main contribution is training a second `high-level' agent which can issue language commands to this `low-level' agent (Fig \ref{fig:schematic}B). These language commands act as sub-goals in a hierarchical setup and this hierarchy allows us to achieve longer and more complex tasks than would be possible without it. 

\textbf{High-level agent}: The `high-level' (HL) agent policy is trained with a supervised training objective, to match the language commands in the data, as well as a reinforcement learning objective to optimize the language commands for goal-directed behavior. 

\textit{Supervised training loss}: Rather than imitating the motor behavior of the expert trajectories from the `Solver' data, we learn a policy to output language commands produced by the `Setter' (which we then use as language input for the LL agent). This data consists of linguistic subgoals $\mathbf{g}_{\leq T} \equiv (\mathbf{g}_0, \mathbf{g}_1, \mathbf{g}_2, \dots, \mathbf{g}_T)$ as well the observations $\mathbf{o}_{\leq T} \equiv (\mathbf{o}_0, \mathbf{o}_1, \mathbf{o}_2, \dots, \mathbf{o}_T)$.  We optimize a policy (using a behavioral cloning, or BC loss) to produce the subgoals $\mathbf{g}_{\leq T} \equiv (\mathbf{g}_0, \mathbf{g}_1, \mathbf{g}_2, \dots, \mathbf{g}_T)$ conditional on the observations $\mathbf{o}_{\leq T} \equiv (\mathbf{o}_0, \mathbf{o}_1, \mathbf{o}_2, \dots, \mathbf{o}_T)$.

\begin{align*}
\mathcal{L}^\textsc{HL}_{BC} = -\frac{1}{B} \sum_{n=1}^{B} \sum_{t=0}^{K} \ln \pi(\mathbf{g}_{n,t} \mid \mathbf{o}_{n,\leq t}),
\end{align*} 

\textit{Reinforcement learning loss}: We can generate simulated trajectories in the tasks specified, where we sample subgoals from the HL agent and issue these to the frozen LL agent. The HL agent is only queried once every 8 timesteps, so the LL agent sees the same language input for 8 timesteps. We then get environment rewards $\mathbf{r}_{\leq T} \equiv (\mathbf{r}_0, \mathbf{r}_1, \mathbf{r}_2, \dots, \mathbf{r}_T)$ from the environment. The LL agent is effectively treated a `part of the environment' for RL training of the HL agent. We used V-trace \citep{espeholt2018} and augment the agent architecture with a value head and optimize:
\begin{align*}
\mathcal{L}^\textsc{HL}_{RL} & = \frac{1}{B} \sum_{n=1}^{B} \sum_{t=0}^{K} \mathbf{R}_{n,t} - \mathbf{V}_{n,t} \ln  \pi(\mathbf{g}_{n,t} \mid \mathbf{o}_{n,\leq t})
\end{align*} 
where $\mathbf{R}_{n,t}$ is total return and $\mathbf{V}_{n,t}$ is the estimated state-value.

\textit{Combining the losses:} The total loss for the HL agent weights the two losses. We vary the relative weights of these losses in experiments.

\begin{align*}
\mathcal{L}^\textsc{HL} & =
w_{BC} \mathcal{L}^\textsc{HL}_{BC} + w_{RL} \mathcal{L}^\textsc{HL}_{RL}
\end{align*}

\section{Experiments}

In all our experiments, we jointly train on all 4 tasks (details in App \ref{sec:task_details}). 
Before pursuing quantitative controls, we first qualitatively describe the behavior observed. As depicted in Figure \ref{fig:schematic}C, we see that the commands generated by the `high level' agent are semantically meaningful. Since the the LL agent is frozen, the high and low level agents cannot develop a different communication protocol via RL -- the HL agent is restricted to using commands that the LL agent trained on human generated instructions can understand. This adds interpretability to the agent's behavior.  

In the next few sections, we ablate the two main factors essential to our approach: the architecture (hierarchical vs flat; Section \ref{sec:hierarchy}) and the loss (cloning expert behavior vs reinforcement from the environment; Section \ref{sec:BCRL}). We also run some analyses on the trained hierarchical agent to understand its behavior (Section \ref{sec:analysis}).

\subsection{Hierarchical agent outperforms flat agent}
\label{sec:hierarchy}
In this section we compare our approach to a flat again that directly produces the actions without a hierarchy. For this baseline, we use the same architecture we used for the HL agent, but change the way it interacts with the environment. Instead of producing language instructions once every 8 timesteps (which are fed to a pre-trained LL agent which outputs environment actions), it directly output environment actions at every timestep to get RL reward. This action head also receives a BC loss computed on the expert environment actions taken by the `Solver'. Recent work shows that simply predicting language sub-goals via an auxiliary head for the LL agent might also help it learn complex tasks by shaping its representations \citep{lampinen2022tell, kumar2022using}. Following these, we also give the agent an auxiliary prediction loss on the language instructions produced by the `Setter' at every time step (see App \ref{sec:app_langweight} for details). This agent has access to all the same data as the hierarchical one, and the architecture is the main difference.

We see in Fig \ref{fig:results}A that the flat agent can pick up on the simpler tasks, but not on the harder tasks -- while the hierarchical agent can learn both. The hierarchical agent also learns the easy tasks faster.

\subsection{Both BC and RL losses are necessary}
\label{sec:BCRL}

\begin{wrapfigure}{l}{0.45\textwidth}
    \centering
    \includegraphics[width=0.9\linewidth]{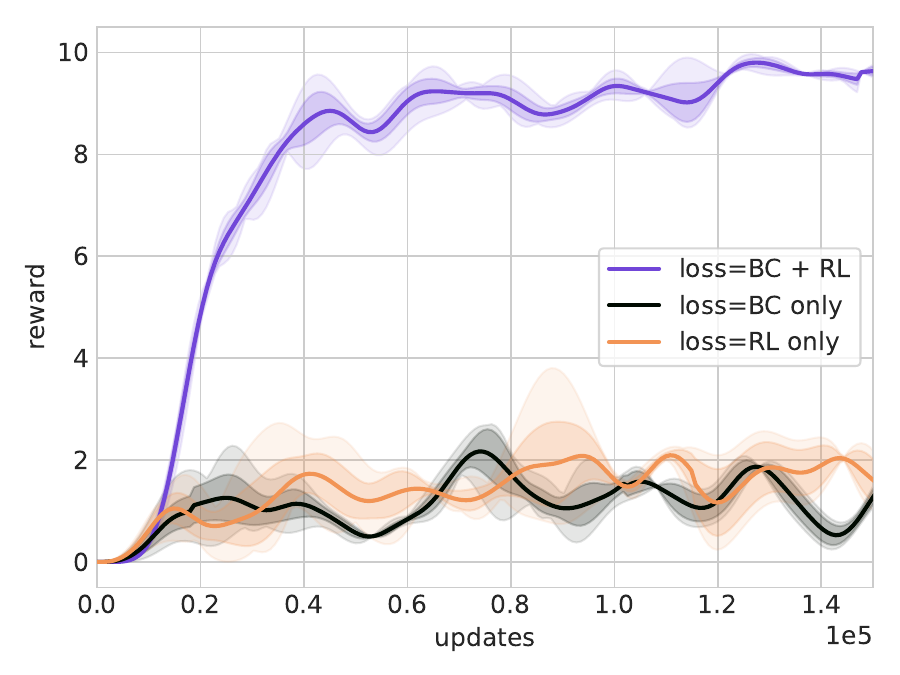}
    \caption{BC and RL losses are needed.}
    \vspace{-10pt}
    \label{fig:BCRL}
\end{wrapfigure}

We now evaluate whether both losses are necessary for the hierarchical agent. We show in Fig \ref{fig:BCRL} that when trained with only BC or only RL loss, the agent cannot learn any of these tasks (within the number of updates we ran for), while the agent trained with both losses learns quickly.

We also split this up by task in the Appendix (Figure \ref{fig:bcrl_long}), this shows that the combination of both losses is necessary to perform reliably well on all the tasks.

\begin{figure*}[t]
    \centering
    \includegraphics[width = 1.0\linewidth]{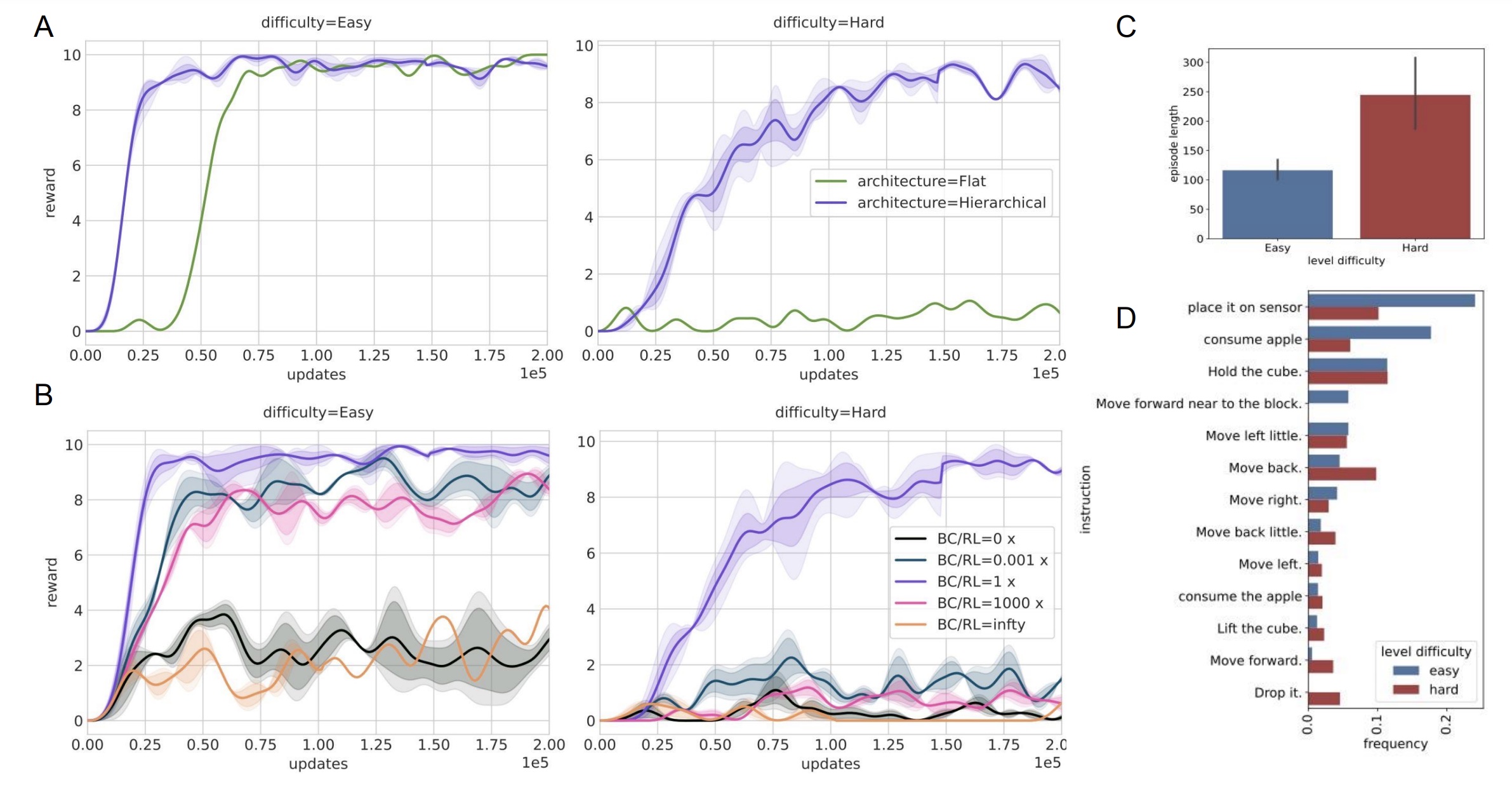}
    \caption{\textbf{Results.} This plot shows results from various sections in the text. A: a hierarchical agent learns much better than a flat agent; details in Section \ref{sec:hierarchy}. B: both BC and RL losses are necessary for good performance; details in Section \ref{sec:BCRL}. C \&  D: analysis of BC+RL hierarchical agent's outputs; details in Section \ref{sec:analysis}. }
    \label{fig:results}
\end{figure*}

We investigate this further by interpolating the relative weights of the BC and RL losses in Figure \ref{fig:results}B. BC/RL = 0 means BC weight is set to zero, and BC/RL = infty means RL weight is set to zero. In addition, we examine the effect of significantly skewing the weights in a 1000:1 ratio both ways.\footnote{Note that the actual value of this relative weight is hard to compute since the losses don't operate over the same scales; we instead vary their relative weights.} We find that significantly over weighting one or the other loss leads to poorer performance, and the best performance comes from placing comparable weight on both losses. As long as we have some weight on each of the two losses (BC and RL), the agent is able to do reasonably well on the Easy tasks, but not on the Hard tasks.
\newline

\subsection{Analysis of the hierarchical BC+RL agent}
\label{sec:analysis}

We then performed some simple analyses of the instructions produced by the hierarchical BC+RL agent. First, we compare the number of instructions required to solve the level, restricting only to levels that were successfully completed. We find that the  Hard levels (that could not be solved by the flat baseline or baselines with skewed BC / RL relative losses) do in fact require many more instructions from the HL agent to complete (Figure \ref{fig:results}C).

We then examined the instructions actually produced. We discarded instructions that appeared less than 100 times across 80 recorded episodes (20 per level). We found that the diversity of instructions required by the hard tasks is higher than that required by the easy tasks -- as noted by the flatter distribution over instructions seen for the red bars in Figure \ref{fig:results}D compared to the blue bars. Further, we note that the instructions that occur more frequently in the hard levels than the easy ones might be associated with error-correcting for the LL agent, most notably the `Drop it' instruction. Generic `Move' commands, notably the `Move back' / `Move forward' command, are also used significantly more in the Hard tasks -- possibly because moving around the level is useful for the information gathering required to find the right key in these tasks.

\section{Discussion and Future Work}
This work contributes to a growing field of language in embodied agents \citep{luketina2019survey}. Closely related is learning in text-based RL games \citep{he2015deep, narasimhan2015language}; we operate in embodied action spaces, and our HL agent handles problems specific to this setting, like error correcting a learned LL agent (Figure \ref{fig:results}D). Language has previously been used as the abstraction in HRL. These use templated language \citep{andreas2017modular, jiang2019language}, or focus on generalization in the LL agent \citep{chen2020ask}. We instead train the HL agent with output supervision from natural language. There are several ways to expand this work. Following several recent papers \citep{ahn2022i, driess2023palme, dasgupta2023collaborating, wang2023voyager}, future work can use pre-trained language models to get a good prior over possible language commands.

\bibliography{main}

\appendix

\newpage

\begin{figure*}[t!]
\begin{center}
\centerline{\includegraphics[width=0.6\linewidth]{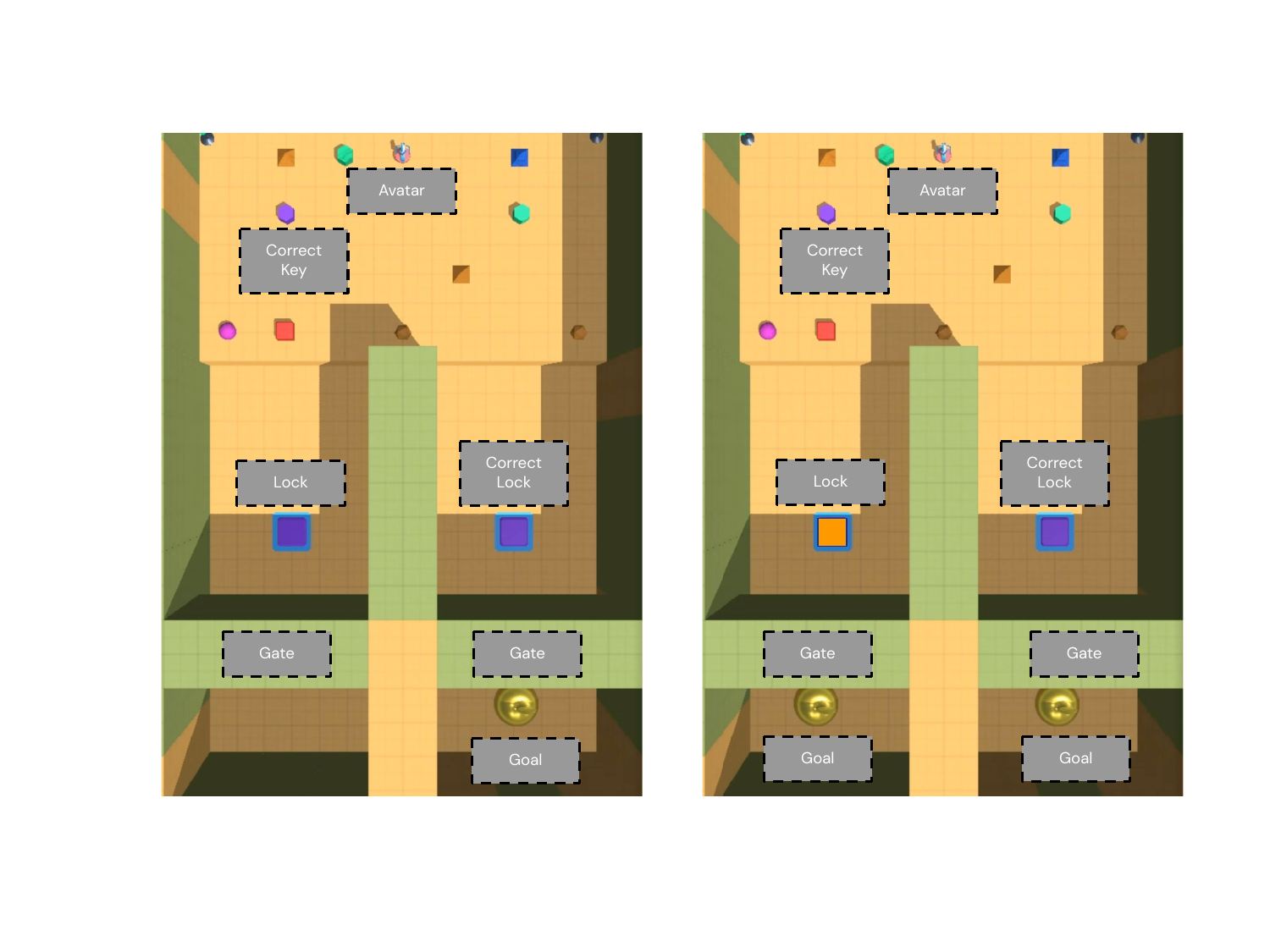}}
\vspace{-1.2cm}
\caption{\textbf{Levels}: A top down view of the two Hard levels: Key Choice Hard (left) and Key+Gate Choice Hard (right). The avatar's initial starting position is also shown. The Easy versions of the tasks have the same structure, but do not contain any distractor keys / objects -- instead the only object / key in the level is the ``correct key" and this is always spawned in front of the avatar's initial position.}
\label{fig:levels}
\end{center}
\end{figure*}

\section{Task details}
\label{sec:task_details}

In all tasks, the agent's goal is to get a goal object (always a golden apple) that is behind a gate. There are two gates (left gate and right gate). Gates can be opened by activating its sensor. A sensor is activated by placing an object (henceforth a `key') with the matching color on it. There are one or more objects accessible to the agent to use as keys -- however the sensor and gate are on a different level as these keys. So the agent has to choose the right key before committing to a gate, it can't just try all possible keys. This induces an information gathering challenge -- the agent has to first investigate which gate a) has an apple behind it and b) is possible to open given the keys available. This information gathering and the multiple steps needed to get the right key and go to the right gate, especially over a fairly high-dimensional action space, make these tasks challenging for traditional RL approaches. The details of the tasks are below.

\subsection{Key Choice}

\begin{itemize}
    \item Choosing the right gate [easy] : The agent can see both gates from the starting position, and only one of them has a goal object behind it. So it knows which gate to target.
    \item Finding the right key [easy] : There is only one key in the level, always placed in the same position. So it knows which object to pick up.
\end{itemize}

\subsection{Key Choice Hard}

\begin{itemize}
    \item Choosing the right gate [easy] : The agent can see both gates from the starting position, and only one of them has a goal object behind it. So it knows which gate to target.
    \item Finding the right key [hard] : There are many objects available to the agent as keys. It has to choose the right key, based on investigating the color of the sensor (which also changes every episode and is not visible from the starting poistion).
\end{itemize}

\subsection{Key+Gate Choice}

\begin{itemize}
    \item Choosing the right gate [hard-er*] : Both gates have a goal object behind them. But only one of them is ``open-able" -- i.e. they both need different keys and only one of the keys is available in the level.
    \item Finding the right key [easy] : There is only one key in the level, always placed in the same position. So it knows which object to pick up.
    
    *Since there is only one available key in the level, the agent's choice of which gate is also made easier (go to the gate with the color matching the only available key) -- rather than having to look at several keys and deciding which one of the gates is open-able.
\end{itemize}

\subsection{Key+Gate Choice Hard}

\begin{itemize}
    \item Choosing the right gate [hard] : Both gates have a goal object behind them. But only one of them is ``open-able" -- i.e. they both need different keys and only one of the keys is available in the level.
    \item Finding the right key [hard] : There are many objects available to the agent as keys. It has to choose the right key, based on which of them matches one of the two possible sensor colors (which change every episode and are not visible from the starting position).
\end{itemize}

\section{Flat agent details}

\label{sec:app_langweight}

\begin{figure*}[t!]
\begin{center}
\centerline{\includegraphics[width=0.8\linewidth]{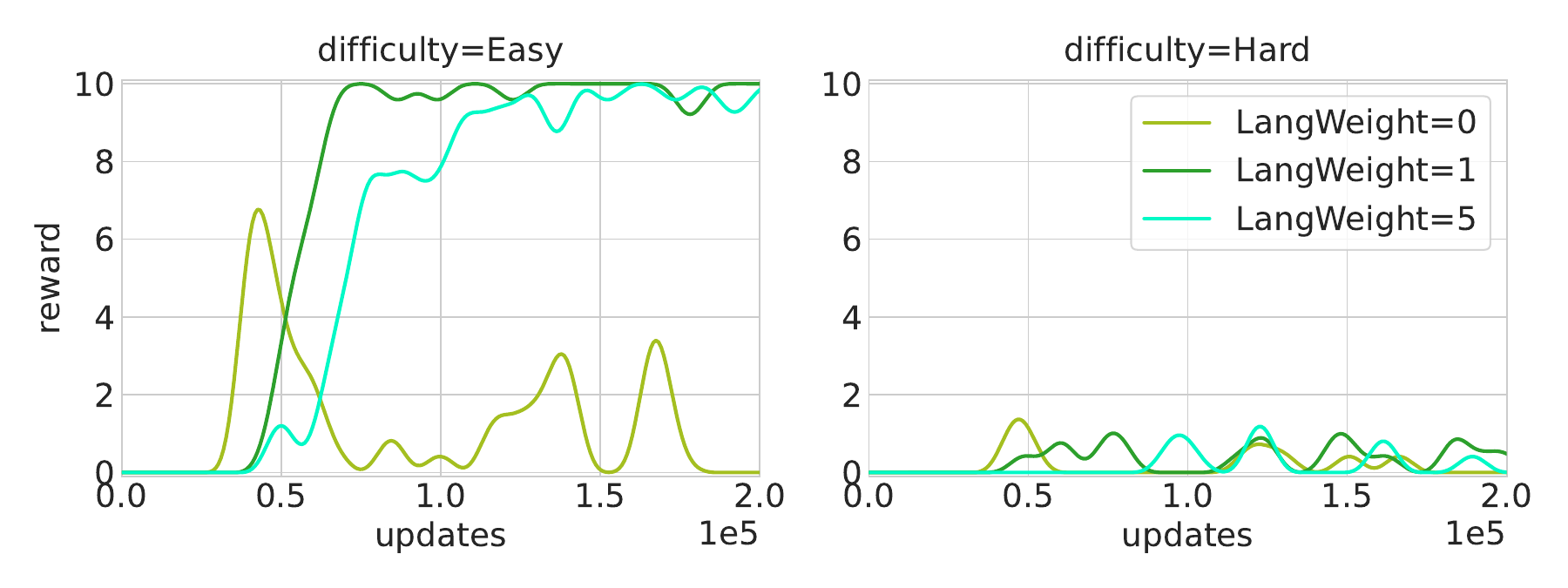}}

\caption{Changing the auxiliary language weight either doesn't affect or reduces performance.}
\label{fig:app_langweight}
\end{center}
\end{figure*}
We tried a few different weights for the auxiliary loss to ensure that the poor performance of our flat agent doesn't stem from a bad value of this parameter. We tried setting this to 0, 1 and 5, and report the results for lang weight = 1 in the main text. Again, the absolute value of this loss is not directly comparable to the RL loss or the loss from predicting the expert actions -- since they don't operate on the same scale. But we find overall that increasing (lang weight = 5) or decreasing this weight (lang weight = 0) either has no effect or reduces performance compared to the setting we report in the main text (lang weight = 1); see Figure \ref{fig:app_langweight}.

\begin{figure*}[htbp]
\centering
   \begin{subfigure}{}
   \includegraphics[width=1.0\textwidth]{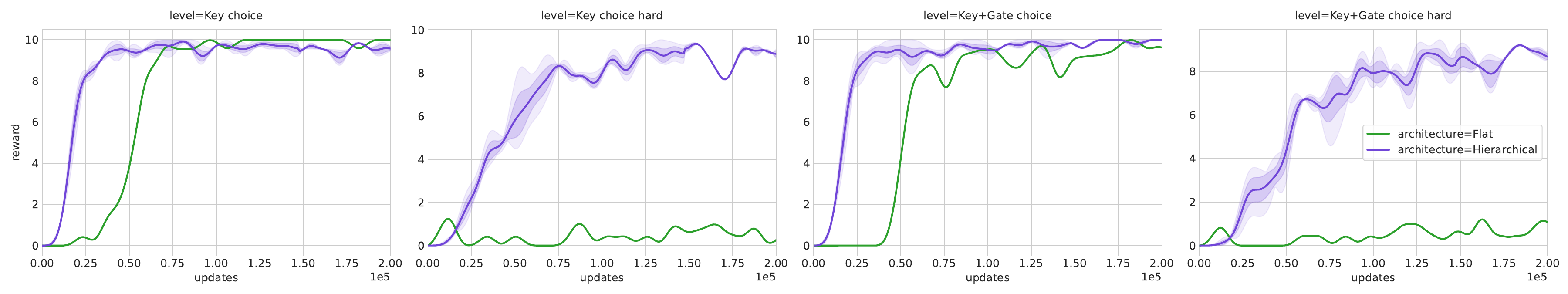}
  \caption{Hierarchical agent does better than a flat agent; Figure \ref{fig:results}A split by level.}
   \label{fig:flat_v_hier_long} 
\end{subfigure}
\begin{subfigure}{}
   \includegraphics[width=1.0\textwidth]{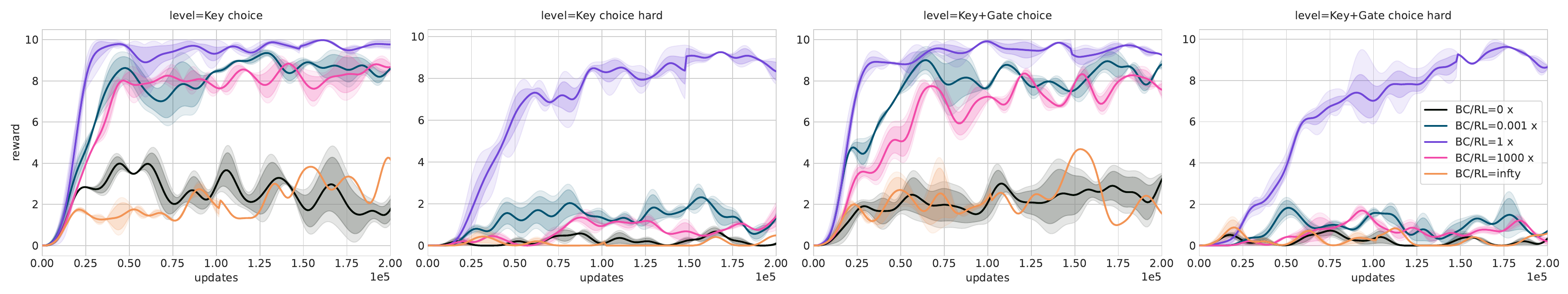}
  \caption{Both BC and RL losses are needed for good performance; Figure \ref{fig:results}B split by level.}
   \label{fig:bcrl_long}
\end{subfigure}
% \caption{}
\end{figure*}

\end{document}